\documentclass{llncs}

\usepackage{amsmath}
\usepackage{amssymb}
\usepackage{bm}
\usepackage{graphicx}
\usepackage{hyperref}
\usepackage[misc]{ifsym}
\usepackage{multirow}
\usepackage{tikz}
\usepackage{url}
\usepackage{xcolor}

\newtheorem{defn}{Definition}

\definecolor{lime}{HTML}{A6CE39}
\DeclareRobustCommand{\orcidicon}{%
	\begin{tikzpicture}
	\draw[lime, fill=lime] (0,0) 
	circle [radius=0.16] 
	node[white] {{\fontfamily{qag}\selectfont \tiny ID}};
	\draw[white, fill=white] (-0.0625,0.095) 
	circle [radius=0.007];
	\end{tikzpicture}
	\hspace{-2mm}
}

\foreach \x in {A, ..., Z}{%
	\expandafter\xdef\csname orcid\x\endcsname{\noexpand\href{https://orcid.org/\csname orcidauthor\x\endcsname}{\noexpand\orcidicon}}
}

\begin{document}
\title{Rule-Guided Graph Neural Networks for Recommender Systems}
%
%
\author{
	Xinze Lyu \and 
	Guangyao Li \and
	Jiacheng Huang\and
	Wei Hu$^\text{(\Letter)}$
}
%
\institute{State Key Laboratory for Novel Software Technology,\\ Nanjing University, Nanjing, China\\
\email{xinzelyu@outlook.com},
\email{\{gyli,jchuang\}.nju@gmail.com},\\
\email{whu@nju.edu.cn}
}
\maketitle              
\begin{abstract}
To alleviate the cold start problem caused by collaborative filtering in recommender systems, knowledge graphs (KGs) are increasingly employed by many methods as auxiliary resources. However, existing work incorporated with KGs cannot capture the explicit long-range semantics between users and items meanwhile consider various connectivity between items. In this paper, we propose RGRec, which combines rule learning and graph neural networks (GNNs) for recommendation. RGRec first maps items to corresponding entities in KGs and adds users as new entities. Then, it automatically learns rules to model the explicit long-range semantics, and captures the connectivity between entities by aggregation to better encode various information. We show the effectiveness of RGRec on three real-world datasets. Particularly, the combination of rule learning and GNNs achieves substantial improvement compared to methods only using either of them.

\end{abstract}

\section{Introduction}

Recommender systems play an important role in modern society. They provide users convenient access to the needed resources out of massive information on the Internet. For services offering content to users like YouTube~\cite{youtube} and Alibaba~\cite{alibaba}, recommender systems are almost a necessity. Collaborative filtering is a widely-used and effective solution, which recommends items by exploring existing user-item interactions. However, collaborative filtering often suffers from the so-called \emph{cold start} problem. It may perform poorly for recommending brand new items or suggesting items to new users. To alleviate this problem, many efforts \cite{cd,tem} have been devoted to designing methods for using auxiliary resources like user or item profiles. In recent years, knowledge graphs (KGs) are increasingly selected. KGs contain structural data of high quality, which provide a wealth of relations between items. Thus, brand new items, which are rarely interacted with users, can be better recommended by the relations recorded in KGs. 

Existing works incorporated with KGs can be roughly divided into three categories: embedding-based, path-based and aggregation-based. The embedding-based methods \cite{cke} often model the direct relations between entities only; they lack the capability of capturing the long-range semantics between entities. A few path-based methods \cite{per} leverage experts to manually construct (meta)paths between users and items, while others \cite{rkge,kprn} learn rules automatically but ignore various relations between different entities; they only consider the relations presented in rules. Aggregation-based methods \cite{kgcn,kgat} model relations between different entities by the attention mechanism. They can preserve rich information around a central entity (i.e. the entity that we want to obtain its representation) by aggregating the representations of its directly or indirectly connected entities. However, it is usually hard to model the explicit relations between the central entity and its indirectly connected entities. Thus, the explicit long-range semantics is still not fully explored in aggregation-based methods.

\begin{figure}[!t]
\centering
\includegraphics[scale=0.35]{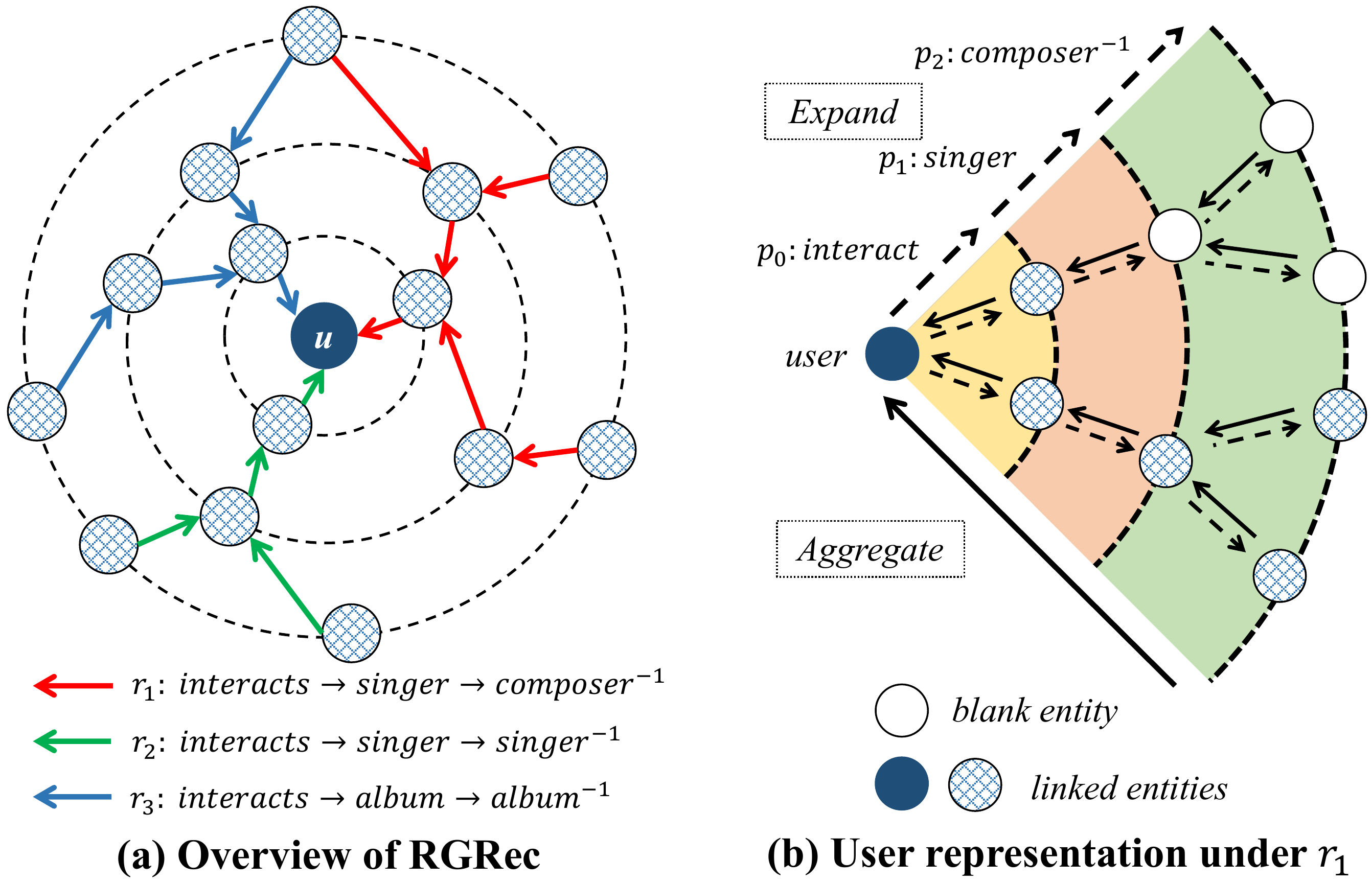}
\caption{Overview of RGRec. Expansion is denoted by dashed arrows, which means that we search the connected entities of $user$ based on each rule. Aggregation is denoted by solid arrows, which means that we combine the representations of $user$ and the entities connected to it.}
\label{fig:combined}
\end{figure}

In this paper, we design RGRec, a method integrating automatic rule learning and graph neural network (GNN)-based aggregation for recommendation. As shown in Figure~\ref{fig:combined}(a), we model the users, items, and entities by a graph, where rules present relation paths between them. Taking $u$ as an example, at first, we extract the entities for $u$ along a rule. Then, we aggregate the representations of entities in the relation path to form the representation of $u$, which can be regarded as the representation on one dimension. We repeat the step on different rules, and form the representations of $u$ on multiple dimensions which corresponds to different rules. Furthermore, different rules have different strengths to extract entities, which corresponds to rules with different confidence. So, the generated multi-dimensional representations are gathered selectively to construct the final representation of $u$. Through this procedure, the rules capture the explicit long-range semantics between entities, and the aggregation makes different entities share their information. Compared with the three categories aforementioned, our method has three key merits:
\begin{enumerate}
\item We combine rule learning and GNNs to capture the long-range semantics between users and items and the connectivity between entities simultaneously. To construct user representations, the rules capture the long-range semantics between users and items, and also guide the procedure of sampling entities, which can alleviate the information loss caused by random sampling in aggregation-based methods. GNNs preserve various connectivity between entities, which can provide richer information to users in addition to rules.

\item We propose strategies to leverage KG embeddings for rule filtering, which provides a more precise way to calculate the confidence of rules when the KGs are incomplete. We also use rule learning techniques to pre-train the weights of rules, which make different rules have different contributions according to their importance.

\item We conduct experiments on three real-world datasets and compare with a number of methods in all the three categories mentioned above. Our results demonstrate the effectiveness of the combination of rule learning and GNNs.
\end{enumerate}

\section{Related Work}

Recommender systems incorporated with KGs can be generally classified into three categories. The first category borrows the idea from KG embedding. MKR \cite{mkr} designs a cross-and-compress unit to share latent features between items in the recommendation task and entities in the KG embedding task. CKE \cite{cke} generates embeddings for structural knowledge with TransR \cite{transr} and combines the embeddings of structural, textual and visual knowledge for collaborative filtering. DKN \cite{dkn} incorporates KG embeddings into news recommendation. It designs a multi-channel and word-entity-aligned knowledge-aware convolutional neural network that fuses word-level and knowledge-level representations of news. These works only consider the direct relations between entities, so they cannot model the long-range semantics between entities.

The second category is based on paths. A part of works uses metapaths, which are defined as the sequences of entity types between users and items, e.g., $user\rightarrow song \rightarrow singer \rightarrow song$. PER \cite{per} introduces metapath-based latent features to represent the connectivity between users and items along different types of paths. It defines recommendation models at both global (same for all users) and personalized levels. FMG~\cite{fmg} incorporates more complex semantics between users and items by introducing metagraphs, which are composed of many different metapaths. HERec~\cite{herec} and metapath2vec~\cite{metapath2vec} use metapaths to sample entities and generate embeddings. MEIRec~\cite{meirec} presents the metapath-guided neighbors to aggregate rich neighbor information. It needs users, items, and queries (a.k.a. intents) as input, and studies the intent recommendation problem, which means that the recommendation for a user is personalized queries rather than items. The performance of metapath-based methods depends heavily on the quality of handcrafted metapaths. To resolve this problem, several works like RKGE~\cite{rkge} and KPRN~\cite{kprn} mine paths (rules) automatically. Although they can capture the long-range semantics between users and items, their strategies to use rules can be improved. For all rules about a user-item pair, the released code of RKGE\footnote{\url{https://github.com/sunzhuntu/Recurrent-Knowledge-Graph-Embedding}} and KPRN\footnote{\url{https://github.com/eBay/KPRN}} shows that they only sample a very small amount of rules randomly. These strategies may omit much useful information. We think that a better way is to delete low-quality rules and save high-quality rules by designing rule filtering algorithms. Generally speaking, modeling with rules is precise because the information is collected by the control of predicates presented in rules, but this also makes the rule-based methods weak in capturing the various connectivity between entities and insufficient in generalization ability.

The third category is characterized by iterative aggregation. RippleNet~\cite{ripplenet} classifies the entities around one entity as $1$-hop, $2$-hop, $\ldots$, $k$-hop neighbors, and aggregates the representations of all these neighbors from different hops in a weighted manner. Differently, KGCN~\cite{kgcn} and KGAT~\cite{kgat} are inspired by GNN architectures like GCN~\cite{gcn}, GraphSage~\cite{graphsage}, GAT~\cite{gat} and HAN~\cite{han} to aggregate the representations of only $1$-hop neighbors around one entity, and the entity will get the information of $k$-hop neighbors by repeating the aggregation $k$ times. Note that, different GNN architectures are designed to capture the information of a graph more precisely, and they are often evaluated on the classification and clustering tasks; while KGCN and KGAT just utilize GNNs to build recommender systems. In these methods, when we choose neighbors for a central entity, we usually cannot know the explicit relations between the central entity and its indirectly connected neighbors. So, less informative neighbors may be collected as noises. Contrary to the path-based methods, the aggregation-based methods are strong in generalization ability because they can capture various connectivity between entities, but weak in precision because the quality of sample entities cannot be guaranteed.

\section{Problem Formulation}

In this paper, we define a KG $\mathcal{G}$ as a set of RDF triples. An RDF triple, denoted by $(s,p,o)$, consists of three components: subject $s$, predicate $p$ and object $o$. According to the common recommendation scenario, we refer to subjects and objects in $\mathcal{G}$ as entities, and the set of entities is denoted by $ \mathcal{E}=\{e_0,\ldots,e_{n_e}\}$. Predicates represent the relations between entities, and the set of predicates is denoted by $\mathcal{P}=\{p_1,\ldots,p_{n_p}\}$.

A typical recommender system contains a set of users $\mathcal{U}=\{u_1,\ldots,u_{n_u}\}$, a set of items $\mathcal{M}=\{m_1,\ldots,m_{n_m}\}$, and the interactions between them (usually modeled as an interaction matrix $\mathcal{H}$). To link $\mathcal{U}$ and $\mathcal{M}$ to KGs, we map an item $m$ in $\mathcal{M}$ to a corresponding entity $e$ in $\mathcal{E}$, then a new triple $( u, interacts, e )$ is added in $\mathcal{G}$, where $u$ is regarded as a new entity and the relation between $u$ and $e$ is denoted by $interacts$. This newly-supplemented $\mathcal{G}$ is denoted by $\mathcal{G}_{\mathcal{H}}$.

A rule specifically refers to an inference rule of predicate $interacts$. So, the rules are means to reason whether a user-item pair instantiates predicate $interacts$. We define the set of rules as $\mathcal{R}=\{r_1,\ldots,r_{n_r}\}$, where a rule $r$ in $\mathcal{R}$ is composed by a set of predicates $\{p, p_1, \ldots, p_h\}$, written as $r:p\Leftarrow p_1 \land \ldots \land p_h $. $p$ on the left of $\Leftarrow$ is called rule head, the part on the right of $\Leftarrow$ is called rule body, and the number of predicates in the rule body is the rule length. When a user $u$ and an item $m$ instantiate a rule $r$, it means that there are entities $\{e_1,e_2,\ldots,e_{h-1}\}$ connecting $u$ and $m$ as $u \stackrel{p_1} {\longrightarrow} e_1 \stackrel{p_2}{\longrightarrow} \ldots \stackrel{p_{h-1}} {\longrightarrow} e_{h-1} \stackrel{p_{h}} {\longrightarrow} m$, which is recorded as $(u,r,m)\in \mathcal{G}_{\mathcal{H}}$. We distinguish directly connected entities ($\text{rule length}=1$) and indirectly connected entities ($\text{rule length}\ge 2$). We believe that rules of length over 1 can help reflect the explicit relations between those indirectly connected entities.

\begin{defn}[Problem definition] Given a KG $\mathcal{G}$ and the interaction matrix $\mathcal{H}$ between users $\mathcal{U}$ and items $\mathcal{M}$, our goal is to learn a function $\mathcal{F}(u,m\,|\,\Theta,\mathcal{R},\mathcal{G}_{\mathcal{H}})$ that can predict the probability of each user-item pair $(u,m)$ instantiating predicate $interacts$, where $u\in\mathcal{U}, m\in\mathcal{M}, \Theta$ denotes the parameter to learn, $\mathcal{R}$ is the set of rules and $\mathcal{G}_{\mathcal{H}}$ is the KG supplemented with $\mathcal{H}$.
\end{defn}

\section{RGRec}

RGRec imitates the ways that humans recommend things and focuses on expressing user features precisely and completely. Taking songs for example, we may consider several aspects when we want to recommend songs to a user $u$. Assume that $u$ likes song $m_1$. We may consider songs that are composed by the singer of $m_1$, or have the same singer as $m_1$,  or are recorded in the same album as $m_1$. These three linear modes of thinking can be expressed by $r_1$, $r_2$ and $r_3$,  respectively:
\begin{small}
\begin{align}
\label{r1}
r_1 : interacts(u,m_1) \Leftarrow &\ interacts(u, m_2) \land singer(m_2,c_1)  \land composer^{-1}(c_1,m_1), \\
\label{r2}
r_2 : interacts(u,m_1) \Leftarrow &\  interacts(u,m_2) \land singer(m_2,s_1)   \land  singer^{-1}(s_1,m_1), \\
\label{r3}
r_3: interacts(u,m_1) \Leftarrow &\ interacts(u,m_2) \land album(m_2,a_1)  \land album^{-1}(a_1,m_1),
\end{align}
\end{small}where $p^{-1}$ denotes the inverse predicate of $p$, e.g., $composer^{-1}(c_1,m_1)$ expresses the same meaning as $composer(m_1,c_1)$.

KGs contain various entities and rich connections, which provide a wealth of resources to generate the representations of users. To construct a user representation, we leverage the rules that can capture the long-range semantics between entities as the guidance. Different rules lead to different user representations, which can be regarded as the representations from various dimensions. A complete user representation is formed by aggregating the collected representations selectively. In Figure~\ref{fig:combined}(a), user $u$ is expanded with three rules, and the expanded entities converge to $u$ in the opposite direction (from outside to inside) iteratively to generate the representation of $u$. To achieve this, we face three challenges:
\begin{enumerate}
\item How to learn rules of high quality?
\item How to model the user representation with a single rule?
\item How to aggregate various representations collected under different rules?
\end{enumerate}

We describe our method in detail in the rest of this section.

\subsection{Rule Learning}
In this paper, we aim to find high-quality inference rules for  predicate $interacts$, which express users like some things. We divide our  rule learning process into two steps: rule finding and rule filtering.

\begin{figure}[!t]
\centering
\includegraphics[scale=0.4]{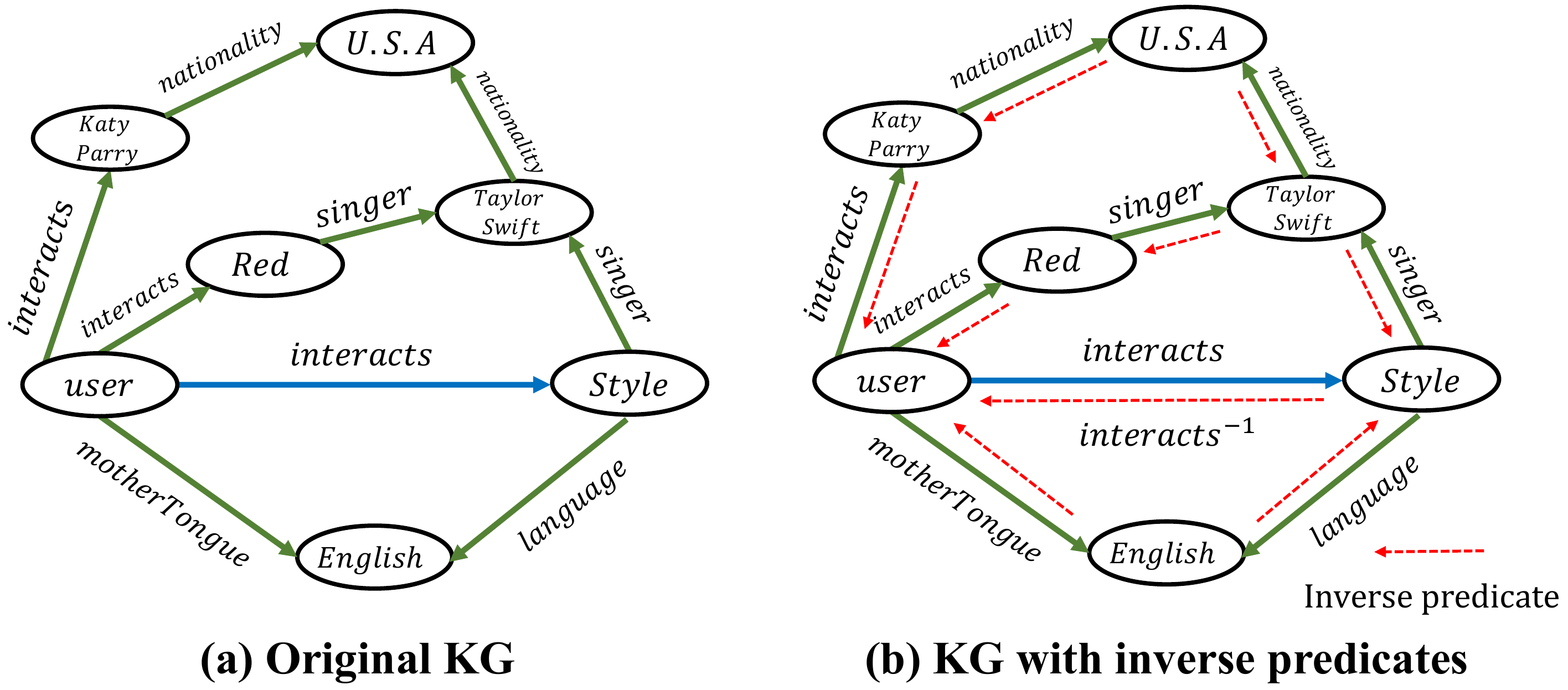}
\caption{A KG fragment}
\label{fig:inverse_predicate}
\end{figure}

For \textbf{rule finding}, we define that each candidate inference rule of $interacts$ is a connected path from a user to an item, where the user and the item instantiate predicate $interacts$ and the direction of predicates in the path is omitted. For example, in Figure~\ref{fig:inverse_predicate}(a), a user interacts with a song called \textit{Style}, the connected paths between the user and \textit{Style} can be regarded as candidate rules. These rules can represent the reasons why this user likes \textit{Style}. For instance, we may infer that the user likes \textit{Style} because the singer of \textit{Style} is the same as a song interacted with the user, through $user \xrightarrow{interacts} Red \xrightarrow{singer} Taylor\  Swift\xrightarrow{singer^{-1}} Style$.

To facilitate path finding, we add an inverse predicate to every edge in the KG like Figure~\ref{fig:inverse_predicate}(b) to make the KG undirected, i.e. adding an inverse triple $(o,p^{-1},s)$ in the KG for every $(s,p,o)$. With the triple $(s,interacts,o)$ as input, we use bidirectional breadth-first search to find connected paths between $s$ and $o$ of length at most $I$ as the candidate rules of $interacts$.

For \textbf{rule filtering}, there are two reasons to adopt it:
\begin{enumerate}
\item Getting rid of low-quality rules that are harmful. In Figure~\ref{fig:inverse_predicate}(b), in addition to the path passing \textit{Red}, two other paths between the user and \textit{Style} are: (1) The mother tongue of the user is English, \textit{Style} is an English song, so the user interacts with \textit{Style} through $user \xrightarrow{motherTongue} English \xrightarrow{language^{-1}} Style$; (2) The user interacts with the singer whose nationality is \textit{U.S.A.}, so the user interacts with other singers from U.S.A. through $user \xrightarrow{interacts} Katy\ Parry \xrightarrow{nationality} U.S.A. \xrightarrow{nationality^{-1}} Taylor\ Swift \xrightarrow{singer^{-1}} Style$. We argue that these two rules are less rational, so rule filtering is necessary.

\item From the implementation aspect, too many rules (e.g., more than 10,000) would challenge the method to keep efficient. Therefore, the number of rules needs to be reduced by filtering for this reason.
\end{enumerate}

As demonstrated in AMIE~\cite{amie}, partial completeness assumption (PCA) and closed world assumption (CWA) are two effective ways to calculate the confidence of rules. CWA assumes that KGs are complete. PCA holds the idea that, if a KG knows some $p$-facts of subject $s$, i.e. the triples involving predicate $p$ of $s$, then it knows all $p$-facts of $s$. So, it neglects the inferred $(s,o)$ whose $s$ is not involved in any $p$-facts. Since users interact with at least one item in our scenario, PCA is identical to CWA for predicate $interacts$. Also, $interacts$ is assumed to be very incomplete in recommendation tasks, i.e. there are many potential items that may interact with users. Consequently, the confidence calculated under CWA may have a great loss. On the other hand, the embeddings of a KG have the ability to complete the graph \cite{transe}. Thus, it can make up the shortcomings of CWA. We design an efficient algorithm to filter rules based on a KG embedding model called RotatE~\cite{rotate}. Below, we briefly describe it. For a triple $(s,p,o)$, RotatE maps $s,p$ and $o$ into a complex vector space and defines $p$ as the rotation from $s$ to $o$. It expects $\bm{o} = \bm{s}\circ \bm{p}$, where $\bm{s},\bm{p},\bm{o}\in \mathbb{C}^{d_{re}}$ denote the embeddings, $\circ$ is the Hadamard (a.k.a. element-wise) product, and the modulus of each element of $\bm{p}$ is 1. RotatE can infer the composition pattern of predicates, e.g., $\bm{p} =\bm{p_1} \circ \ldots\circ \bm{p_h}$ holds if the rule $r : p \ \Leftarrow p_1\land\ldots\land p_h$ is absolutely correct. So, the distance between $\bm{p_1} \circ \ldots\circ \bm{p_h}$ and $\bm{p}$ can reflect the confidence of $r$, which is calculated as follows:
\begin{align}
\label{rule_conf}
conf(r)&\  = -||\,\bm{p} - f(\bm{p})\,||_2, \\
\label{r_eq}
f(\bm{p})&\ =\,\bm{p_1} \circ \ldots\circ \bm{p_h}, 
\end{align}
where $||\cdot||_2$ represents the $L_2$-norm of a complex vector. Taking $r_1$ in Eq.~(\ref{r1}) as an example, we denote $interacts, singer, composer^{-1}$ by $p_t, p_s$ and $p_c^{-1}$, respectively. If $r_1$ is correct, i.e.  $\bm{m_1} =  \bm{u} \circ \bm{p_t}, \bm{m_2} = \bm{u}\circ \bm{p_t}, \bm{c_1} = \bm{m_2}\circ \bm{p_s}$ and $\bm{m_1} = \bm{c_1}\circ \bm{p_c^{-1}}$ hold, $\bm{p_t}  = \bm{p_t} \circ \bm{p_s} \circ \bm{p_c^{-1}}$ can be inferred by $\bm{u} \circ \bm{p_t} = \bm{c_1} \circ \bm{p_c^{-1}} = (\bm{m_2}\circ \bm{p_s}) \circ \bm{p_c^{-1}} = ((\bm{u}\circ \bm{p_t})\circ \bm{p_s})\circ \bm{p_c^{-1}} $. 
  
Finally, we reserve top-$L$ rules with the highest confidence as output. 

In addition to RotatE, DistMult~\cite{dismult} and RLvLR~\cite{rlvlr} can also use the composition pattern of predicates and embeddings to measure the confidence of rules. However, DistMult, which represents relations by matrices in bilinear transformation, needs special constraints to infer the composition pattern of predicates, but the constraints may not hold in implementation. RotatE points out that DistMult cannot infer the composition pattern of predicates, but TransE~\cite{transe} and itself can \cite{rotate}. For RLvLR, by only using the composition pattern, it performs poorly when rules are longer than 2. So, it designs another strategy based on co-occurrence for longer rules. Compared with these two methods, our strategy of using the embeddings trained by RotatE to model the composition pattern of predicates is theoretically reasonable and practical in reality. Still, RotatE has some detrimental effects, such as the fixed composition pattern mentioned in QuatE~\cite{quate}. This causes the performance of RotatE not particularly good when some predicates participating in a compositional pattern are the same. We will consider other advanced KG embedding models to alleviate this problem in future work.

\subsection{User Representation Guided by Single Rule}

Inspired by GraphSAGE~\cite{graphsage}, which is a general inductive framework for graph representation learning, we design a rule-guided GNN model. To learn the representation of a user $u$ along a rule $r$, we firstly select fixed-size $k$-hop neighbors of $u$ along $r$. Then, we aggregate the representations of entities to their directly connected neighbors and apply a non-linear transformation to construct the representations of entities aware of neighbors. Finally, we repeat this process for a few iterations to make $u$ receive the information from all selected neighbors. We take rule $r_1$ in Eq.~(\ref{r1}) as an example to explain how to obtain the representation of a user under the guidance of $r_1$. As shown in Figure~\ref{fig:combined}(b), we expand the user along the rule (direction: $Expand$), then we aggregate the representations of the expanded entities to the user reversely (direction: $Aggregate$) to obtain the representation of the user under this rule.

We define the $k$-hop expanding entity set of user $u$ on $r$ as $\mathcal{D}_u^k(r) = \{ o \,|\,  (s, p_k,$ $o),s \in \mathcal{D}_u^{k-1}(r)\}$, where $k\in [1,h]$ and $\mathcal{D}_u^0(r)=\{u\}$. When we expand $u$ along $r$, if there exist entities in $\mathcal{D}_u^{k-1}(r)$ that cannot use the $k$-th predicate to conduct the $k$-hop expansion, then it receives a negative feedback that $r$ is infeasible for $u$ to some extent. In practice, we return a blank entity $B$ as the negative feedback for $\{s \,|\,\nexists\, o, (s,p_k,o) \in \mathcal{G}_{\mathcal{H}}, s \in \mathcal{D}_u^{k-1}\}$. The blank entities are shown as white circles in Figure~\ref{fig:combined}(b). 

\begin{figure}[!t]
\centering
\includegraphics[width=.85\columnwidth]{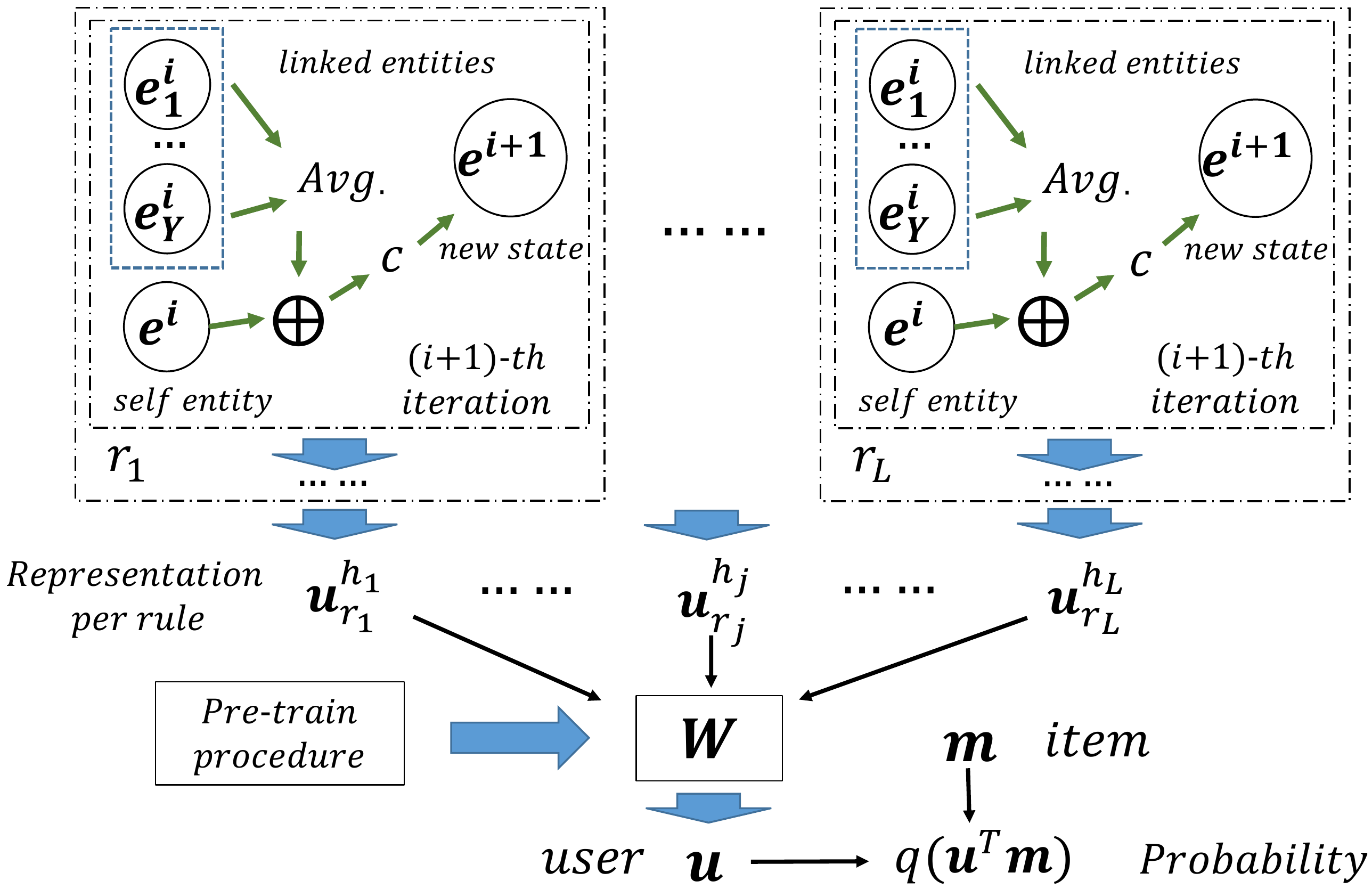}
\caption{The framework of RGRec}
\label{fig:whole_model}
\end{figure}

The entity set whose representations to be aggregated is denoted by $\mathcal{J}_i = \{\mathcal{D}_u^0(r)\cup \ldots\cup \mathcal{D}_u^{h-i}(r)\}$, where $i\in[1,h]$, and $h$ is the length of rule $r$, which is also the total number of aggregation iterations in $r$. The aggregation proceeds from $\mathcal{J}_1$ to $\mathcal{J}_h$ in turn. The $(i+1)$-th iteration is shown in the upper part of Figure~\ref{fig:whole_model}. During this iteration, the state $\bm{e}$ of entity $e$  (self entity) in $\mathcal{J}_{i+1}$ is transformed from $\bm{e}^{i}$ to $\bm{e}^{i+1}$ (new state) as follows:
\begin{align}
\label{update_i_iteration}
\bm{e}^{i+1} &= c\Big( \bm{e}^i \oplus (\frac{1}{Y}\sum_{y=1}^{Y}\bm{e}_y^i) \Big), \\
\label{specific_func_update_i_iteration}
c(\bm{x}) &=\sigma(\bm{W}_{agg}\bm{x}+\bm{b}),
\end{align}
where the states of entities which should be aggregated to $\bm{e}^i$ are denoted by $\{ \bm{e}_1^i, \ldots, \bm{e}_{Y}^i\}$ (linked entities), $\bm{e}$ is of size $d_r$, $\oplus$ means vector concatenation, and $\sigma$ is a nonlinear function like $Sigmoid$. At each round of iterations, RGRec applies the aggregation operation to entities along the direction of $Expand$, where $user$ is the first entity to be applied the aggregation operation in the first iteration. After $h$ iterations, the final representation of $u$ under $r$ is $\bm{u}_r^h$.

\subsection{Multi-dimensional Representation Aggregation}
\label{subsection_Multi-dimensional Representation Aggregation}

Given $L$ representations $\{\bm{u}_{r_1}^{h_1}, \bm{u}_{r_2}^{h_2},\ldots,\bm{u}_{r_{L}}^{h_L}\}$ of user $u$ under the guidance of $L$ rules $\{r_1,r_2,\ldots,r_{L}\}$, where $h_j$ denotes the length of $r_j$, the final representation $\bm{u}$ of $u$ is aggregated as follows: 
\begin{align}
\label{final_representation}
\bm{u}=\left[ \bm{u}_{r_1}^{h_1}; \bm{u}_{r_2}^{h_2}; \ldots; \bm{u}_{r_{L}}^{h_L} \right] \bm{W},
\end{align}
where $\bm{W}$ is the rule weights of size $(L\times 1)$ and the size of $\left[ \bm{u}_{r_1}^{h_1}; \bm{u}_{r_2}^{h_2}; \ldots; \bm{u}_{r_{L}}^{h_L} \right]$ is $(d_r \times L)$.

The loss function $loss_\text{RGRec}$ of RGRec is defined as
\begin{align}
\label{loss_RGRec}
loss_\text{RGRec}=\frac{1}{N}\sum_{i=1}^N\Big(l_i-q(\bm{u}_i^T\bm{m}_i)\Big)^2+\mu ||\bm{W}||_2,
\end{align}
where, for $N$ training data $\{(\bm{u}_i,\bm{m}_i,l_i)\}_{i=1}^{N}$, $\bm{u}_i$, $\bm{m}_i$ and $l_i$ are the user representation, item representation and label (1 if the user and the item instantiate predicate $interacts$, and 0 otherwise), respectively. $\mu$ is the hyperparameter of $L_2$-regularization. $\bm{m}_i$ has size $(d_r \times 1)$. $q$ is a nonlinear function like $Sigmoid$. The idea behind the loss function is that, if a user $u$ and an item $m$ instantiate $interacts$, their label $l$ is 1, and the inner product of their representations is expected to be 1; otherwise, their label is 0, and the inner product is expected to be 0. $\mu ||\bm{W}||_2$ is a regularization term to avoid overfitting.

\subsection{Rule Weights Pre-training}
\label{subsection_Rule Weights Pre-training}

Not every rule should play an equal role  during the formation of the final representation. However, the confidence of rules calculated by embeddings does not work well here. That confidence only measures whether the rules can interpret predicate $interacts$. It checks rules in isolation and lacks the consideration for the whole rule set.  In fact, rules can affect each other, including both positive  and negative influences. For example, if $r_1$ (Eq.~(\ref{r1})) or $r_3$ (Eq.~(\ref{r3})) hold between  user $u$ and  item $m$, $m$ is less likely to be recommended to $u$ just by one rule, but when $m$ has the same composer and belongs to the same album as one song that $u$ interacts, i.e. $r_1$ and $r_3$ both hold, the probability of being recommended is higher. In this paper, we design a pre-training procedure to learn rule weights $\bm{W}$ automatically from a more holistic perspective.

Assume that we have $L$ rules $\mathcal{R} = \{r_1, r_2, \ldots, r_{L}\}$ for $interacts$ and $N$ user-item pairs $\{(u_i,m_i)\}_{i=1}^{N}$. For each user-item pair that instantiates $interacts$, we label it with 1, otherwise we label it with 0. The label set for all user-item pairs is denoted by $\{l_i\}_{i=1}^N$. Then, we test every user-item pair $(u_i,m_i)$ against every rule $r_j$, i.e. returning 1 if  $(u_i,r_j,m_i) \in \mathcal{G}_{\mathcal{H}}$, and 0 otherwise, which generates the feature set $\{\bm{X}_i\}_{i=1}^N$. $\bm{X}_i$ is a vector of size $L$ and composed of $0/1$.

With training data $\{(\bm{X}_i,l_i)\}_{i=1}^{N}$, we convert the problem whether the user and the item instantiate $interacts$ to a binary classification problem, and the parameters to learn are the rule weights $\bm{W}$. The loss function is defined as
\begin{align}
\label{loss_pra}
loss_\text{pre-train}=\frac{1}{N}\sum_{i=1}^N\Big(l_i-z(\bm{W}^T\bm{X}_i)\Big)^2+\lambda ||\bm{W}||_2,
\end{align}
where $\lambda$ is the hyperparameter of $L_2$-regularization and $z$ is a nonlinear function like $Sigmoid$. $\bm{W}$ is pre-trained in Eq.~(\ref{loss_pra}) and fine-tuned in Eq.~(\ref{loss_RGRec}) to obtain the representations of users and items. Here, our method to form each feature vector $\bm{X}_i$ is inspired by PRA~\cite{pra}. Each dimension feature is corresponding to the probability of the connectivity between the user and the item by the relation path. We simplify the process by assigning 0/1 to each feature, which makes the procedure more efficient.

\section{Experiments and Results}

We implement RGRec on a workstation with an Intel Core i9-9900K CPU, 64 GB memory and a NVIDIA GeForce RTX 2080 Ti graphics card. The source code is available online\footnote{\url{https://github.com/nju-websoft/RGRec}}. In our experiments, we want to answer the following two research questions:
\begin{enumerate}
\item[Q1.] Compared to the state-of-the-art rule-based and GNN-based methods, how does RGRec perform? Are rule learning and GNNs both effective? Particularly, does RGRec work well in the cold start scenario?

\item[Q2.] How do rule length and number, rule filtering strategy, and rule weights pre-training affect the overall performance? 
\end{enumerate}

\subsection{Preparation}

\textbf{Datasets.} We pick three real-world datasets: Last.FM (released in KGCN~\cite{kgcn}), MovieLens-1M (in RippleNet~\cite{ripplenet}) and Dianping-Food (in KGCN-LS~\cite{kgcnls}). They all use Microsoft Satori\footnote{\url{https://searchengineland.com/library/bing/bing-satori}} to prepare the corresponding KGs. The statistical data of the three datasets are depicted in Table~\ref{tab:dataset}, where ``\#Entities", ``\#Predicates" and ``\#KG triples" denote the numbers before complementing interaction matrix $\mathcal{H}$. Following conventions~\cite{kgcnls,kgcn}, we split all the data to $\text{training} : \text{validation} : \text{testing} = 6:2:2$.

\begin{table}[!t]
\setlength{\tabcolsep}{5pt}
\centering
\caption{Statistical data of the datasets}
\label{tab:dataset}
\begin{tabular}{|l|rrr|}
	\hline & \multicolumn{1}{c}{Last.FM} & \multicolumn{1}{c}{MovieLens-1M} & \multicolumn{1}{c|}{Dianping-Food} \\
	\hline	\#Users		& 1,872		& 6,036		& 2,298,698	\\
			\#Items		& 3,864		& 2,445		& 1,362		\\
			\#Interactions	& 42,346		& 753,772		& 23,416,418	\\
	\hline	\#Entities		& 9,366		& 182,011		& 28,115		\\
			\#Predicates	& 60			& 12			& 7			\\
			\#KG triples	& 15,518		& 1,241,995	& 160,519		\\
	\hline
\end{tabular}
\end{table}

\medskip
\noindent\textbf{Evaluation metrics.} We use two sets of metrics: AUC and F1 under the click through rate scenario, and Hits@$k$ and NDCG@$k$ ($k\in\{5,10\}$) under the top-$k$ recommendation scenario. To reduce the complexity of measuring Hits@$k$ and NDCG@$k$ during the testing stage, following KPRN~\cite{kprn}, we sample 100 negatives for one positive. Also, following KGCN~\cite{kgcn}, we implement AUC and F1 with the ratio of positives and negatives being $1:1$. Each experiment is repeated five times and the average results are reported. 

\medskip
\noindent\textbf{Hyperparameters.}
For RotatE, we select the implementation in~\cite{openke}. The dimension of predicate embeddings $d_{re}$ is set to 1,024, and other hyperparameters strictly follow the settings in~\cite{openke}. For RGRec, we perform a grid search. The used hyperparameters are determined by optimizing AUC on the validation set with the early stop strategy, i.e. stopped if not improved in successive three epochs. As a result, we set the maximum length of rules $I=3$, the maximum number of used rules $L=30$, the dimension of entity embeddings $d_r=8$, the number of neighbors for every entity $Y=4$, the learning rate to 0.05 for Last.FM and to 0.0005 for MovieLens-1M and Dianping-Food, the $L_2$-regularization parameter $\mu=0.0001$, and the batch size to 128 for Last.FM and to 64 for MovieLens-1M and Dianping-Food. To pre-train rule weights $\bm{W}$, we assign the $L_2$-regularization parameter $\lambda=0.0001$, the learning rate to 0.0001 and the batch size to 256. For the choices of non-linear functions $q,z$ and $\sigma$, we set $q$ and $z$ to $Sigmoid$, and $\sigma$ to $ReLU$ for non-last iterations and to $tanh$ for the last iteration. 

\medskip
\noindent\textbf{Competitors.} We pick SVD~\cite{SVD}, LibFM~\cite{libfm}, LibFM+TransE, PER~\cite{per}, RKGE \cite{rkge}, CKE~\cite{cke}, KGCN~\cite{kgcn} and KGAT~\cite{kgat} as our competitors. SVD and LibFM are two classical methods for recommendation. LibFM+TransE adds embeddings trained by TransE~\cite{transe} to LibFM. PER represents those methods using manually constructed metapaths, while RKGE represents those methods mining paths automatically. CKE is a typical embedding-based method. KGAT and KGCN represent the aggregation-based methods. The hyperparameters for the competitors follow the settings in \cite{kgcn} or the settings suggested in their original papers. We develop SVD, LibFM, LibFM+TransE, RKGE and CKE by ourselves, while reuse the source code of KGAT and KGCN. We cannot implement PER because the three datasets do not provide entity types to construct metapaths. The results of PER on Last.FM, MovieLens-1M and Dianping-Food are quoted from~\cite{ripplenet,kgcnls,kgcn}, respectively, and the results of KGAT on Dianping-Food is missing due to the scalability issue.

\subsection{Results and Analysis}

\begin{table}[!t]
\setlength{\tabcolsep}{5pt}
\centering
\caption{AUC and F1 in the click through rate scenario}
\label{tab:auc}
\begin{tabular}{|l|cccccc|}
	\hline & \multicolumn{2}{c}{Last.FM} & \multicolumn{2}{c}{MovieLens-1M} & \multicolumn{2}{c|}{Dianping-Food} \\
	\cline{2-7} & AUC & F1 & AUC & F1 & AUC & F1 \\
	\hline	SVD			& 0.772	& 0.683	& 0.833	& 0.757	& 0.787	& 0.729 	\\
			LibFM		& 0.773	& 0.716	& 0.830	& 0.777	& 0.809	& 0.766	\\
			LibFM+TransE	& 0.726	& 0.669	& 0.825	& 0.772	& 0.820	& 0.761	\\
			PER			& 0.633	& 0.596	& 0.712	& -		& 0.746	& -		\\
			CKE			& 0.727	& 0.649	& 0.771	& 0.680	& 0.773	& 0.703	\\
			RKGE		& 0.745	& 0.689	& 0.894	& 0.825	& 0.847	& 0.766	\\
			KGCN		& 0.797	& 0.719	& 0.869	& 0.789	& 0.842	& 0.774	\\
			KGAT		& 0.706& 0.709	& 0.906	& \textbf{0.838}& -	& -	\\
			RGRec		& \textbf{0.825} & \textbf{0.747} & \textbf{0.913} & \textbf{0.838} & \textbf{0.884} & \textbf{0.809} \\
	\hline
\end{tabular}
\end{table}

Based on our experimental results, we answer the two research questions as follows. For Q1, as illustrated in Tables~\ref{tab:auc}, \ref{tab:hits} and~\ref{tab:ndcg}, RGRec achieves the overall best AUC, F1, Hits@$k$ and NDCG@$k$ ($k\in\{5,10\}$) on all the three datasets, except for NDCG@5 and NDCG@10 on Last.FM. 

Specifically, we find that (1) for the aggregation-based methods, KGAT achieves competitive AUC and F1 on MovieLens-1M, and KGCN is stable and can be seen as the second best competitor. Compared with them, RGRec shows that rules indeed have the power to guide the aggregation of entity representations. (2) For other methods, PER obtains the worst AUC and F1 on all the three datasets, because it heavily relies on the quality of metapaths manually created. This also demonstrates the advantage of RGRec in learning rules automatically. (3) RKGE has poor Hits@$k$ and NDCG@$k$ ($k\in\{5,10\}$) due to the fact that, although RKGE uses rules during training,  it does not use rules during testing. In fact, it only computes the inner product of user embeddings and item embeddings during testing to resolve the complexity of rule searching. RGRec does not have this problem because rules are searched in advance and the search process is only executed once. 

Furthermore, we use 20$\%$, 40$\%$ and 60$\%$ of the data for training to see the performance of RGRec in the cold start scenario. Limited by the space, we only report the results on the largest Dianping-Food dataset, using AUC and F1 as the metrics. The results on the other two datasets using Hits@$k$ and NDCG@$k$ exhibit a similar phenomenon. As depicted in Table~\ref{tab:cold_start}, RGRec obtains the best and stable results when 20$\%$, 40$\%$ and 60$\%$ (i.e. the default setting) of the data for training are used. We can also see that the performance of several competitors (e.g., KGCN) significantly drops with fewer training data. This verifies the capability of RGRec to address the cold start problem.

\begin{table}[!t]
\setlength{\tabcolsep}{5pt}
\centering
\caption{Hits@$k$ ($k\in\{5,10\}$) in the top-$k$ recommendation scenario}
\label{tab:hits}
\begin{tabular}{|l|cccccc|}
	\hline & \multicolumn{2}{c}{Last.FM} & \multicolumn{2}{c}{MovieLens-1M} & \multicolumn{2}{c|}{Dianping-Food} \\
	\cline{2-7} & Hits@5 & Hits@10 & Hits@5 & Hits@10 & Hits@5 & Hits@10 \\
	\hline	SVD			& 0.357	& 0.501	& 0.306	& 0.511	& 0.384	& 0.557	\\
			LibFM		& 0.396	& 0.539	& 0.304	& 0.513	& 0.380	& 0.582	\\
			LibFM+TransE	& 0.344	& 0.453	& 0.234	& 0.438	& 0.355	& 0.542	\\
			CKE			& 0.188	& 0.294	& 0.070	& 0.134	& 0.351	& 0.526	\\
			RKGE		& 0.058	& 0.122	& 0.152	& 0.251	& 0.090	& 0.167	\\
			KGCN		& 0.417	& 0.551	& 0.333	& 0.537	& 0.295	& 0.479	\\
			KGAT		& 0.284	& 0.394	& 0.235& 0.340	& -& -	\\
			RGRec		& \textbf{0.450} & \textbf{0.571} & \textbf{0.394} & \textbf{0.562} & \textbf{0.43} & \textbf{0.606} \\
	\hline
\end{tabular}
\end{table}%
\begin{table}[!t]
\setlength{\tabcolsep}{5pt}
\centering
\caption{NDCG@$k$ ($k\in\{5,10\}$) in the top-$k$ recommendation scenario}
\label{tab:ndcg}
\resizebox{\columnwidth}{!}{
\begin{tabular}{|l|cccccc|}
	\hline & \multicolumn{2}{c}{Last.FM} & \multicolumn{2}{c}{MovieLens-1M} & \multicolumn{2}{c|}{Dianping-Food} \\
	\cline{2-7} & NDCG@5 & NDCG@10 & NDCG@5 & NDCG@10 & NDCG@5 & NDCG@10 \\
	\hline	SVD			& 0.240	& 0.287	& 0.186	& 0.252	& 0.249	& 0.305	\\
			LibFM		& 0.267	& 0.313	& 0.183	& 0.250	& 0.238	& 0.303	\\
			LibFM+TransE	& 0.244	& 0.279	& 0.137	& 0.203	& 0.233	& 0.293	\\
			CKE			& 0.122	& 0.156	& 0.042	& 0.063	& 0.231	& 0.288	\\
			RKGE		& 0.033	& 0.053	& 0.095	& 0.126	& 0.054	& 0.079	\\
			KGCN		& \textbf{0.325}	& \textbf{0.373}	& 0.236	& 0.306	& 0.216	& 0.279	\\
			KGAT		& 0.198	& 0.233	& 0.154	& 0.188	& -	& -	\\
			RGRec		& 0.324	& 0.363	& \textbf{0.271}	& \textbf{0.325}	& \textbf{0.298}	& \textbf{0.354}	\\
	\hline
\end{tabular}}
\end{table}

\begin{table}[!b]
\setlength{\tabcolsep}{5pt}
\centering
\caption{AUC and F1 on Dianping-Food in the cold start scenario}
\label{tab:cold_start}
\begin{tabular}{|l|ccc|ccc|}
	\hline & \multicolumn{3}{c|}{AUC} & \multicolumn{3}{c|}{F1} \\
	\cline{2-7} & 20\% & 40\% & 60\% & 20\% & 40\% & 60\% \\
	\hline	SVD			& 0.709	& 0.762	& 0.787	& 0.648	& 0.704	& 0.729	\\
			LibFM		& 0.812	& 0.814	& 0.809	& 0.761 	& 0.766	& 0.766	\\
			LibFM+TransE	& 0.798	& 0.819	& 0.820	& 0.747	& 0.760	& 0.761	\\
			CKE			& 0.710	& 0.743	& 0.773	& 0.614	& 0.671	& 0.703	\\
			RKGE		& 0.703	& 0.811	& 0.847	& 0.628	& 0.719	& 0.766	\\
			KGCN		& 0.774	& 0.807	& 0.842	& 0.719	& 0.742	& 0.774	\\
			RGRec		& \textbf{0.882} & \textbf{0.884} & \textbf{0.884} & \textbf{0.808} & \textbf{0.809} & \textbf{0.809} \\
	\hline
\end{tabular}
\end{table}

\begin{figure}[!t]
\centering
\includegraphics[scale=0.32]{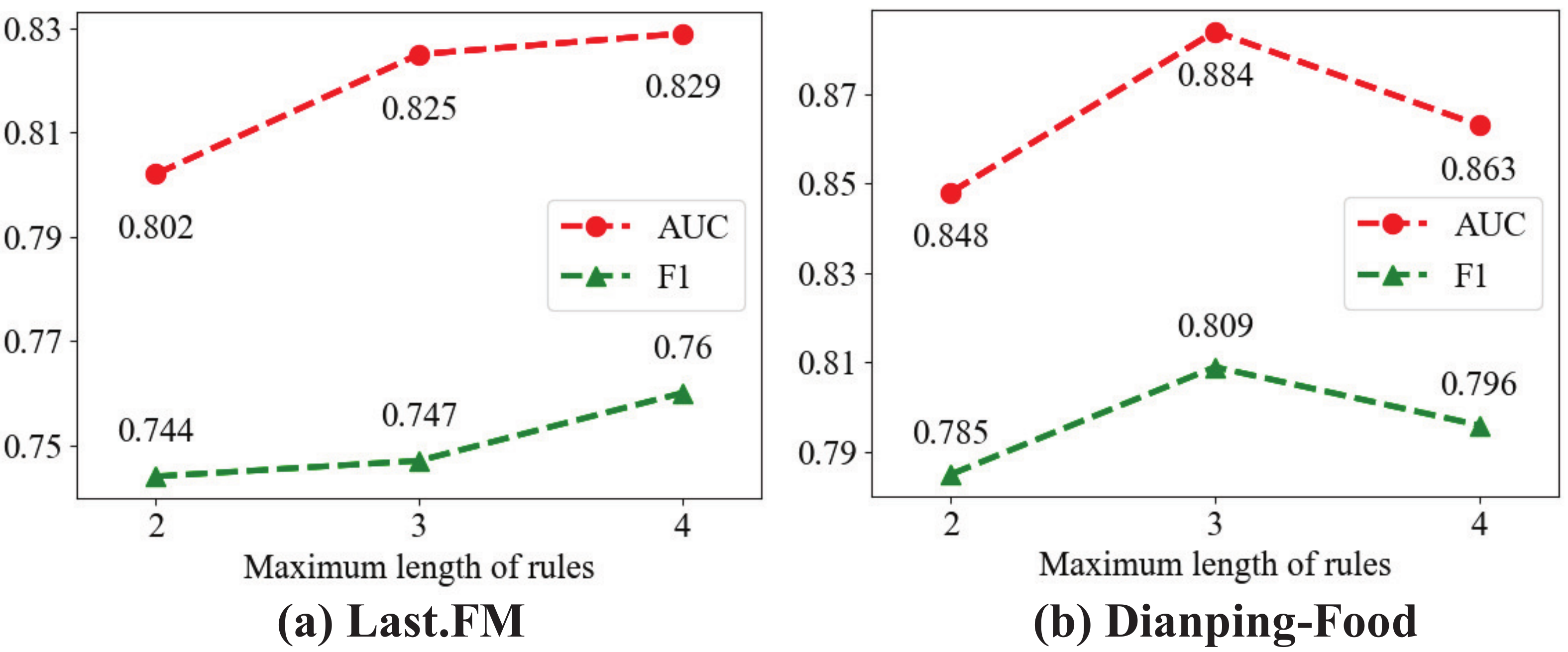}
\caption{AUC and F1 varying with the maximum lengths of rules}
\label{fig:rule_len}
\end{figure}

For Q2, the maximum length of rules is a sensitive parameter. The length of rules indicates the number of iterations for aggregation, which is also called the depth of GNNs in some methods. Deep GNNs can help central entities get information from farther entities but also lead to the over-smoothing problem~\cite{oversmoothing}, i.e. the representations of different entities would become indistinguishable.  Also, in some aggregation-based methods~\cite{ripplenet,kgcn,kgat}, the maximum distance between a central entity and its neighbors is four, which corresponds to rules of length four. Thus, we search the rules of maximum length two, three and four on the three datasets and show the statistics in Table~\ref{tab:rule_len}. Note that, we cannot find the rules of length two and four on MovieLens-1M, so Figure~\ref{fig:rule_len} only shows how the performance of RGRec varies on Last.FM and Dianping-Food. RGRec achieves the best results on Last.FM when the maximum length is four and on Dianping-Food when the maximum length is three.  However, the performance difference is pretty subtle. In practice, we prefer to use three. We believe that this length usually makes sense in recommender systems, like $r_1$ (Eq.~(\ref{r1})), $r_2$ (Eq.~(\ref{r2})) and $r_3$ (Eq.~(\ref{r3})).

\begin{table}[!t]
\setlength{\tabcolsep}{5pt}
\centering
\caption{Number of rules w.r.t. different lengths}
\label{tab:rule_len}
\begin{tabular}{|c|rrr|}
	\hline Lengths & Last.FM & MovieLens-1M & Dianping-Food \\
	\hline	2	& 6		& 0	& 1	\\
			3	& 51		& 54	& 8	\\
			4	& 335	& 0	& 12	\\
	\hline
\end{tabular}
\end{table}

\begin{figure}[!t]
\centering
\includegraphics[scale=0.32]{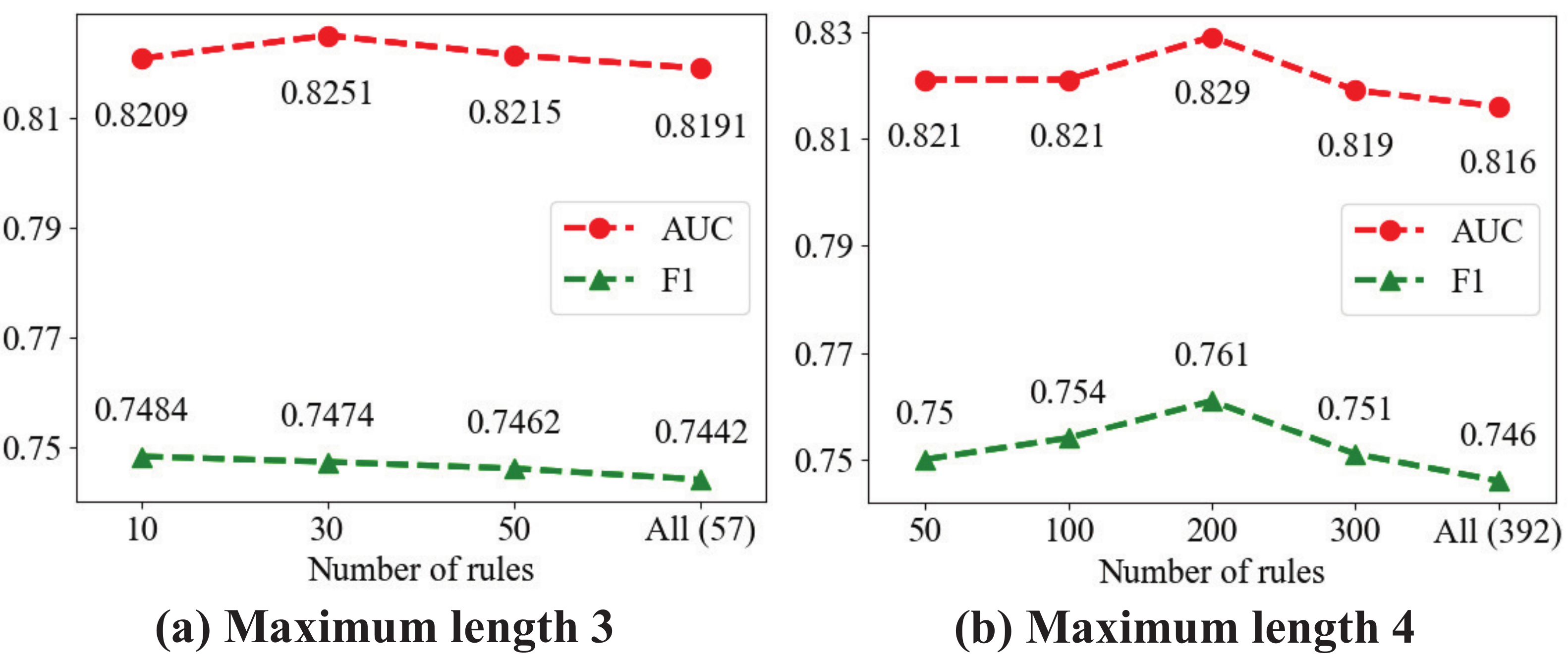}
\caption{AUC and F1 with top-$L$ ranked rules preserved in Last.FM when the maximum lengths of rules are 3 and 4}
\label{fig:rule_num}
\end{figure}

To explore the effect of rule filtering strategies, RGRec is assessed with different numbers of rules preserved in Last.FM when the maximum lengths of rules are 3 and 4. 
MovieLens-1M and Dianping-Food  have much less number of rules 
than Last.FM, so
they are less suitable than Last.FM for this 
experiment.
The results are shown in Figure~\ref{fig:rule_num}.  RGRec does not perform the best when using all rules, which demonstrates that some low-quality rules are harmful and must be eliminated. The strategy of rule filtering succeeds in controlling the quality.

Additionally, we assess four strategies for rule filtering: CWA (closed world assumption), RLvLR~\cite{rlvlr}, TransE~\cite{transe} and RotatE~\cite{rotate}, which are denoted by RGRec$_\text{CWA}$, RGRec$_\text{RLvLR}$, RGRec$_\text{TransE}$ and RGRec$_\text{RotatE}$, respectively. We compare them on Last.FM when the maximum length of rules is 3. We show AUC and F1 with top-$L$ reserved rules in Table~\ref{tab:filtering}. Considering the best results, the highest AUC and F1 of these four methods are not achieved when all rules are used, which verifies the effectiveness of rule filtering. RGRec$_\text{RotatE}$ performs slightly better than the other three, showing that it is more capable of modeling the composition pattern of predicates. Also, embeddings  overcome the incompleteness of KGs to some extent. 

\begin{table}[!t]
\setlength{\tabcolsep}{5pt}
\centering
\caption{AUC and F1 of different filtering strategies on Last.FM}
\label{tab:filtering}
\begin{tabular}{|c|c|cccc|}
	\hline & Top-$L$ & RGRec$_\text{CWA}$ & RGRec$_\text{RLvLR}$ & RGRec$_\text{TransE}$ & RGRec$_\text{RotatE}$ \\
	\hline	\multirow{3}{*}{AUC}		& 10	& 0.8146	& 0.8204	& 0.8127	& \textbf{0.8209}	\\
								& 30	& 0.8179	& 0.8244	& 0.8202	& \textbf{0.8251}	\\
								& 50	& 0.8195	&0.8163	& 0.8141	& \textbf{0.8215}	\\
	\cline{2-6}					& All (57)	& \multicolumn{4}{c|}{0.8191}	\\
	\hline	\multirow{3}{*}{F1}		& 10	& 0.7408	& 0.7451	& 0.7397	& \textbf{0.7484}	\\
								& 30	& 0.7466	& \textbf{0.7479}	& 0.7476	& 0.7474	\\
								& 50	& \textbf{0.7470}	& 0.7419	& 0.7381	& 0.7462	\\
	\cline{2-6}					& All (57)	& \multicolumn{4}{c|}{0.7442}	\\
	\hline
\end{tabular}
\end{table}

\begin{table}[!t]
\setlength{\tabcolsep}{5pt}
\centering
\caption{AUC and F1 of RGRec, RGRec$_{\text{wo}\_\bm{W}}$ and the best competitor}
\label{tab:wo_W}
\begin{tabular}{|l|cccccc|}
	\hline & \multicolumn{2}{c}{Last.FM} & \multicolumn{2}{c}{MovieLens-1M} & \multicolumn{2}{c|}{Dianping-Food} \\
	\cline{2-7} & AUC & F1 & AUC & F1 & AUC & F1 \\
	\hline	Best competitor			& 0.797& 0.719	& 0.906	& \textbf{0.838}& 0.847	& 0.774	\\
			RGRec$_{\text{wo}\_\bm{W}}$		& 0.787	& 0.703	& 0.910	& 0.836	& 0.879	& 0.806	\\
			RGRec					& \textbf{0.825} & \textbf{0.747} & \textbf{0.913} & \textbf{0.838} & \textbf{0.884} & \textbf{0.809} \\
	\hline
\end{tabular}
\end{table}

To explore the effect of rule weights pre-training, we disable the pre-training procedure and build RGRec$_{\text{wo}\_\bm{W}}$. As depicted in Table~\ref{tab:wo_W}, RGRec$_{\text{wo}\_\bm{W}}$ underperforms RGRec on all the three datasets. However, compared with the best competitor, RGRec$_{\text{wo}\_\bm{W}}$ is still competitive on MovieLens-1M and Dianping-Food. We conclude that the pre-training procedure can improve the predictive capability of RGRec.

\section{Conclusion}

In this paper, we propose RGRec, which combines rule learning and GNNs for recommendation. Rules capture the explicit long-range semantics between entities, and GNNs aggregate the information of captured entities along the rules to learn precise representations of users. RGRec achieves superior performance on three real-world datasets. Furthermore, the combination of rule learning and GNNs is better than only using either of them. In future work, we will leverage multi-modal learning to build a more powerful recommender system.\\

\noindent\textbf{Acknowledgments.} This work is supported by the National Natural Science Foundation of China (No. 61872172), the Water Resource Science \& Technology Project of Jiangsu Province (No. 2019046), and the Key R\&D Program of Jiangsu Science and Technology Department (No. BE2018131).

\bibliographystyle{splncs04}
\bibliography{iswc20}

\begin{thebibliography}{10}
\providecommand{\url}[1]{\texttt{#1}}
\providecommand{\urlprefix}{URL }
\providecommand{\doi}[1]{https://doi.org/#1}

\bibitem{cd}
Bayer, I., He, X., Kanagal, B., Rendle, S.: A generic coordinate descent
  framework for learning from implicit feedback. In: WWW. pp. 1341--1350 (2017)

\bibitem{transe}
Bordes, A., Usunier, N., Garcia-Dur{\'{a}}n, A., Weston, J., Yakhnenko, O.:
  Translating embeddings for modeling multi-relational data. In: NIPS. pp.
  2787--2795 (2013)

\bibitem{youtube}
Covington, P., Adams, J., Sargin, E.: Deep neural networks for {YouTube}
  recommendations. In: RecSys. pp. 191--198 (2016)

\bibitem{metapath2vec}
Dong, Y., Chawla, N.V., Swami, A.: metapath2vec: Scalable representation
  learning for heterogeneous networks. In: KDD. pp. 135--144 (2017)

\bibitem{meirec}
Fan, S., Zhu, J., Han, X., Shi, C., Hu, L., Ma, B., Li, Y.: Metapath-guided
  heterogeneous graph neural network for intent recommendation. In: KDD. pp.
  2478--2486 (2019)

\bibitem{amie}
Gal{\'{a}}rraga, L., Teflioudi, C., Hose, K., Suchanek, F.M.: {AMIE}:
  Association rule mining under incomplete evidence in ontological knowledge
  bases. In: WWW. pp. 413--422 (2013)

\bibitem{graphsage}
Hamilton, W.L., Ying, Z., Leskovec, J.: Inductive representation learning on
  large graphs. In: NeurIPS. pp. 1024--1034 (2017)

\bibitem{openke}
Han, X., Cao, S., Lv, X., Lin, Y., Liu, Z., Sun, M., Li, J.: {OpenKE}: An open
  toolkit for knowledge embedding. In: EMNLP. pp. 139--144 (2018)

\bibitem{gcn}
Kipf, T.N., Welling, M.: Semi-supervised classification with graph
  convolutional networks. In: ICLR (2017)

\bibitem{SVD}
Koren, Y.: Factorization meets the neighborhood: A multifaceted collaborative
  filtering model. In: KDD. pp. 426--434 (2008)

\bibitem{pra}
Lao, N., Mitchell, T., Cohen, W.: Random walk inference and learning in a large
  scale knowledge base. In: EMNLP. pp. 529--539 (2011)

\bibitem{oversmoothing}
Li, Q., Han, Z., Wu, X.: Deeper insights into graph convolutional networks for
  semi-supervised learning. In: AAAI. pp. 3538--3545 (2018)

\bibitem{transr}
Lin, Y., Liu, Z., Sun, M., Liu, Y., Zhu, X.: Learning entity and relation
  embeddings for knowledge graph completion. In: AAAI. pp. 2181--2187 (2015)

\bibitem{rlvlr}
Omran, P.G., Wang, K., Wang, Z.: Scalable rule learning via learning
  representation. In: IJCAI. pp. 2149--2155 (2018)

\bibitem{libfm}
Rendle, S.: Factorization machines with {libFM}. ACM Transactions on
  Intelligent Systems and Technology  \textbf{3}(3), ~57 (2012)

\bibitem{herec}
Shi, C., Hu, B., Zhao, W.X., Philip, S.Y.: Heterogeneous information network
  embedding for recommendation. IEEE Transactions on Knowledge and Data
  Engineering  \textbf{31}(2),  357--370 (2018)

\bibitem{rotate}
Sun, Z., Deng, Z.H., Nie, J.Y., Tang, J.: {RotatE}: Knowledge graph embedding
  by relational rotation in complex space. In: ICLR (2019)

\bibitem{rkge}
Sun, Z., Yang, J., Zhang, J., Bozzon, A., Huang, L., Xu, C.: Recurrent
  knowledge graph embedding for effective recommendation. In: RecSys. pp.
  297--305 (2018)

\bibitem{gat}
Velickovic, P., Cucurull, G., Casanova, A., Romero, A., Li{\`{o}}, P., Bengio,
  Y.: Graph attention networks. In: {ICLR} (2018)

\bibitem{ripplenet}
Wang, H., Zhang, F., Wang, J., Zhao, M., Li, W., Xie, X., Guo, M.: {RippleNet}:
  Propagating user preferences on the knowledge graph for recommender systems.
  In: CIKM. pp. 417--426 (2018)

\bibitem{dkn}
Wang, H., Zhang, F., Xie, X., Guo, M.: {DKN}: Deep knowledge-aware network for
  news recommendation. In: WWW. pp. 1835--1844 (2018)

\bibitem{kgcnls}
Wang, H., Zhang, F., Zhang, M., Leskovec, J., Zhao, M., Li, W., Wang, Z.:
  Knowledge-aware graph neural networks with label smoothness regularization
  for recommender systems. In: KDD. pp. 968--977 (2019)

\bibitem{mkr}
Wang, H., Zhang, F., Zhao, M., Li, W., Xie, X., Guo, M.: Multi-task feature
  learning for knowledge graph enhanced recommendation. In: WWW. pp. 2000--2010
  (2019)

\bibitem{kgcn}
Wang, H., Zhao, M., Xie, X., Li, W., Guo, M.: Knowledge graph convolutional
  networks for recommender systems. In: WWW. pp. 3307--3313 (2019)

\bibitem{alibaba}
Wang, J., Huang, P., Zhao, H., Zhang, Z., Zhao, B., Lee, D.L.: Billion-scale
  commodity embedding for e-commerce recommendation in {Alibaba}. In: KDD. pp.
  839--848 (2018)

\bibitem{kgat}
Wang, X., He, X., Cao, Y., Liu, M., Chua, T.: {KGAT}: Knowledge graph attention
  network for recommendation. In: KDD. pp. 950--958 (2019)

\bibitem{tem}
Wang, X., He, X., Feng, F., Nie, L., Chua, T.S.: {TEM}: Tree-enhanced embedding
  model for explainable recommendation. In: WWW. pp. 1543--1552 (2018)

\bibitem{kprn}
Wang, X., Wang, D., Xu, C., He, X., Cao, Y., Chua, T.: Explainable reasoning
  over knowledge graphs for recommendation. In: AAAI. pp. 5329--5336 (2019)

\bibitem{han}
Wang, X., Ji, H., Shi, C., Wang, B., Ye, Y., Cui, P., Yu, P.S.: Heterogeneous
  graph attention network. In: WWW. pp. 2022--2032 (2019)

\bibitem{dismult}
Yang, B., Yih, W., He, X., Gao, J., Deng, L.: Embedding entities and relations
  for learning and inference in knowledge bases. In: ICLR (2015)

\bibitem{per}
Yu, X., Ren, X., Sun, Y., Gu, Q., Sturt, B., Khandelwal, U., Norick, B., Han,
  J.: Personalized entity recommendation: A heterogeneous information network
  approach. In: WSDM. pp. 283--292 (2014)

\bibitem{cke}
Zhang, F., Yuan, N.J., Lian, D., Xie, X., Ma, W.Y.: Collaborative knowledge
  base embedding for recommender systems. In: KDD. pp. 353--362 (2016)

\bibitem{quate}
Zhang, S., Tay, Y., Yao, L., Liu, Q.: Quaternion knowledge graph embeddings.
  In: NeurIPS. pp. 2735--2745 (2019)

\bibitem{fmg}
Zhao, H., Yao, Q., Li, J., Song, Y., Lee, D.L.: Meta-graph based recommendation
  fusion over heterogeneous information networks. In: KDD. pp. 635--644 (2017)

\end{thebibliography}

\end{document}